\documentclass[a4paper, 10pt, conference]{ieeeconf}      

\IEEEoverridecommandlockouts                           
\usepackage{multirow}

\usepackage{amsfonts}
\usepackage{amsmath}
\usepackage{wrapfig}
\usepackage[algo2e]{algorithm2e} 
\usepackage{mathtools}
\usepackage{color} 
\usepackage{float}
\usepackage{subfigure}
\usepackage{tikz}
\usepackage{algorithm}
\usepackage{placeins}
\usepackage{bbm}
\usepackage{hyperref}
\usepackage[noend]{algpseudocode}
\DeclareMathOperator*{\argmax}{argmax}

\usepackage{amsmath,amsfonts,amssymb,graphicx,bm}

\title{\LARGE \bf
Modeling Multi-Vehicle Interaction Scenarios \\ Using Gaussian Random Field
}

\author{Yaohui Guo$^{1}$, Vinay Varma Kalidindi$^{1}$,
Mansur Arief$^{1}$,
Wenshuo Wang$^{1}$, \\
Jiacheng Zhu$^{1}$,
Huei Peng$^{2}$,
Ding Zhao$^{1,*}$
\thanks{The first two authors contributed equally to this work}
\thanks{$^{1}$Department of Mechanical Engineering, Carnegie Mellon University, Pittsburgh, PA 15213, USA}%
\thanks{$^{2}$Department of Mechanical Engineering, University of Michigan, Ann Arbor, MI 48109, USA.}
\thanks{*Corresponding author. E-mail: {\tt\small dingzhao@cmu.edu}}
}

\begin{document}

\maketitle
\thispagestyle{empty}
\pagestyle{empty}


\begin{abstract}
Autonomous vehicles are expected to navigate in complex traffic scenarios with multiple surrounding vehicles. The correlations between road users vary over time, the degree of which, in theory, could be infinitely large, thus posing a great challenge in modeling and predicting the driving environment.
In this paper, we propose a method to model multi-vehicle interactions using a stochastic vector field model and apply non-parametric Bayesian learning to extract the underlying motion patterns from a large quantity of naturalistic traffic data. We then use this model to reproduce the high-dimensional driving scenarios in a finitely tractable form. We use a Gaussian process to model multi-vehicle motion, and a Dirichlet process to assign each observation to a specific scenario. 
We verify the effectiveness of the proposed method on highway and intersection datasets from the NGSIM project, in which complex multi-vehicle interactions are prevalent. The results show that the proposed method can capture motion patterns from both settings, without imposing heroic prior, and hence demonstrate the potential application for a wide array of traffic situations.
The proposed modeling method could enable simulation platforms and other testing methods designed for autonomous vehicle evaluation, to easily model and generate traffic scenarios emulating large scale driving data.
\end{abstract}
\section{Introduction}
\label{sec:introduction}

The deployment of an autonomous vehicle (AV) on public roads requires the AV  to be able to interact with various driving scenarios involving multiple road users. In addition, there is a growing public expectation that an AV shall be able to drive and merge seamlessly in complex traffic \cite{claybrook2018autonomous}. Hence, the modeling of multi-vehicle interaction scenarios is inevitable. 
Traditionally, researchers and engineers in transportation rely on strong assumptions to keep the inference tractable. Researchers often impose some prior knowledge regarding driving scenarios \cite{deo2018would}, consider all road users using the same driving strategies \cite{treiber2000congested}, or simplify the systems by only simulating one-to-one interactions \cite{zhao2017accelerated, wang2018extracting}, which restrict the applicability of such models for the study of naturalistic driving systems. To alleviate these assumptions, the model should be able to consider the interactions among vehicles within dynamic driving scenes, while avoiding the restriction of presupposing the number of vehicles involved, which is a challenging task. In order to fulfil these modeling requirements, we employ Gaussian Process (GP) to represent the multi-vehicle motion model. 

 GP has been proven to be effective in modeling trajectory patterns, especially for prediction and classification. Research in \cite{kim2011gaussian} uses flow fields generated from Gaussian Process Regression Flow (GPRF) to represent motion trajectories and performs classification and prediction based on some prior knowledge regarding the driving situation and the nature of the interaction. \cite{Barao2018} uses a Gaussian vector random field to model observed trajectories and cluster them using k-means algorithm. However, a Gaussian model has rarely been used for modeling multi-vehicle interactions. 

Many existing multi-vehicle modeling methods focus on trajectory prediction. In \cite{deo2018MultiModal}, LSTM (Long Short-Term Memory) is used to model surrounding vehicles and predict their motion with the experiments limited to highway datasets. \cite{MultimodalDCN} uses a deep convolutional neural network to output predicted trajectories with associated probabilities. \cite{JensInteraction} uses context aware, Markovian models to describe multi-agent behavior and dynamic Bayesian networks to perform the prediction. In \cite{QTran}, a Gaussian process regression is applied for multi-modal maneuver recognition and trajectory prediction.  
 
 The main contribution of this study is a fully data-driven framework to model multi-vehicle interactions and is achieved by combining the effectiveness of multi-dimensional GP \cite{GPforML} to model motion patterns and the versatility of the Dirichlet Process (DP) \cite{dirichletprocess} in enabling the data to determine the number of these patterns. The combination of these methods yields a non-parametric method to model the  stochastic system in a Bayesian view, alleviating the need of pre-specifying the scenarios. In modeling motion patterns for trajectory prediction, \cite{joseph2011bayesian} shows DP-GP performs better than Markov-based models, but only with the experiments limited to a single vehicle. DP-GP modeling has also been used in the prediction of pedestrian trajectories \cite{ChenPedestrain}. While prediction could be seen as a popular outcome of modeling multi-agent motion using DP-GP, we also study how it could also prove useful in capturing the interaction scenarios based on traffic data and generate simulated trajectories emulating the data.
 
 Using the proposed framework, we are able to model multi-vehicle interactions of given driving data sequence and extract the underlying scenarios as motion patterns. The motion patterns learned shall fully characterize the observed traffic scenes and for any given (fixed or dynamic) number of vehicles and their initial conditions, could be used to generate driving scenarios representational of real-world traffic data. The ability to generate realistic scenarios based on the learned motion patterns is highly beneficial for AV development, e.g. efficient evaluation. Crude evaluation of AVs could involve millions of miles of testing on road and billions of miles on simulation platforms \cite{waymo2017, GMreport2018}.
 Such a scenario extraction and reconstruction model, combined with an efficient evaluation scheme, e.g. \cite{zhao2017accelerated}, could be used to strategically simulate important cases, hence efficiently testing the competencies of AVs under naturalistic driving scenarios. 
 
\section{Multi-vehicle Motion Model}\label{sec:model}
The proposed multi-vehicle motion model is defined as a mixture $G$ of motion patterns
\begin{equation}\label{eqn_Mixture}
    G = \sum ^{K}_{k=1} \pi_{k} g_{k}
\end{equation}
where each mixture component $g_{k}$ is a motion pattern and is defined by a GP. The respective mixture weights $\pi_{k}$ sum to 1 and are defined using a DP prior. Consider a dataset with N observations, where each observation contains the position and velocity information of all the vehicles in a given region of interest. Each of these N observations shall be assigned to one of the K motion patterns, with each pattern consisting of at least one observation. The problem can be therefore divided into several parts: defining motion patterns $g_{k}$, determining the number of mixtures $K$, and inferring the model parameters.

\subsection{GP Motion Patterns}
\subsubsection{Modeling motion pattern with GP velocity field}
\begin{figure}[h]
	\centering
	\includegraphics[width=\linewidth]{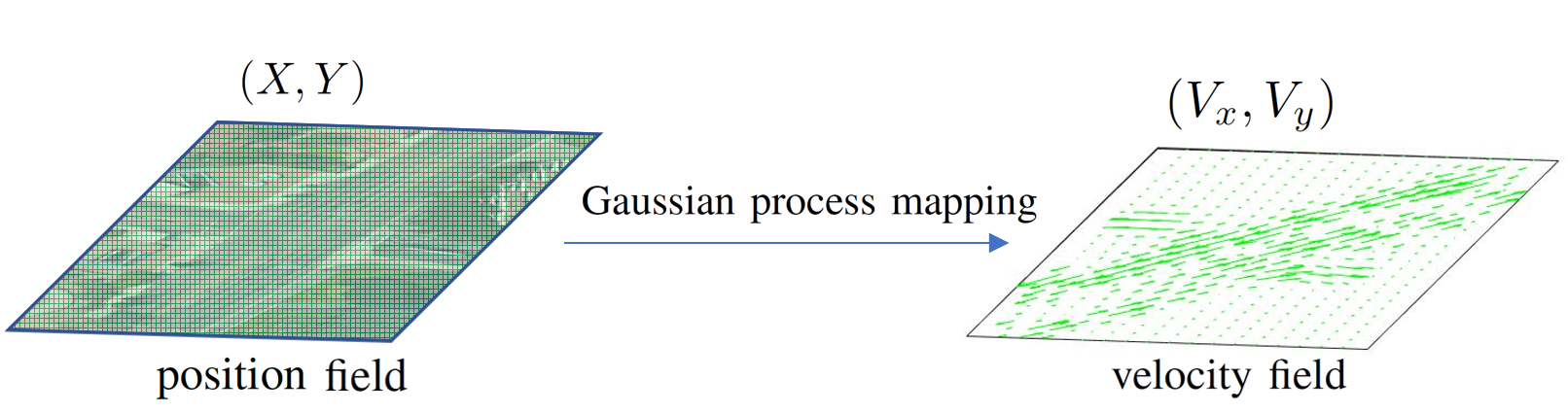}
	\caption{GP mapping from position to velocity vector field in $A_{ROI}$ }
	\label{fig_GP}
\end{figure}

We define a motion pattern $\displaystyle g$ as a GP that maps from position domain to velocity, as illustrated by Fig. \ref{fig_GP}.
\begin{equation}
\begin{aligned}
  g:(x,y)\rightarrow (v_{x} ,v_{y})\\
    \quad (x,y)\in A_{ROI}
\end{aligned}
\end{equation}
where $A_{ROI}$ is the region of interest.

A random field is simply a stochastic process, taking values in a Euclidean space and defined over a parameter space of dimensionality at least one \cite{RandomField}. Therefore, the terms `Gaussian process' and `Gaussian random field' can be used interchangeably here. Given that the current formulation has GP mapping to velocity domain, the Gaussian random field here is effectively the velocity field.

The velocity field information within a small region is expected to be consistent-- a property that motivated the exploitation of GP to capture the consistency. A GP is defined as a collection of random variables, whose arbitrary subset has a Gaussian distribution \cite{GPforML}. A GP motion pattern $\displaystyle g$ models the velocity $(v_{x} ,v_{y})$ as Gaussian random variables. Here, we assume $v_{x}$ and $v_{y}$ are independent for simplicity. For concise representation, we shall use indicator $\eta \in \{x,y\}$ to avoid writing equations for both $x$ and $y$ directions--- for example, the velocity $(v_{x},v_{y})$ at $(x,y)$ is written as $v_{\eta}( x,y)$ or simply, $v_{\eta}$. The GP here is given by
\begin{equation}\label{velGP}
v_{\eta}( x,y) \sim \mathcal{GP}(  \mu _{\eta} (x,y),\ \mathrm{cov}_{\eta}(x,y) )
\end{equation}
with its mean and covariance functions as
\begin{equation*}
\begin{split}
\mu _{\eta} (x,y) =& \ \mathbb{E}[ v_{\eta} (x,y)]\\
\mathrm{cov}_{\eta}(x,y) =& \ \mathbb{E}[( v_{\eta} (x,y)-\mu _{\eta}( x,y))  \\
& \hspace{12mm} ( v_{\eta} (x',y') -\mu _{\eta}( x',y'))] \\
=& \ k_{\eta}(( x,y) ,( x',y')) + \varepsilon
\end{split}
\end{equation*}
where $(x',y')$ is any position coordinate in $A_{ROI}$, $k_{\eta}$ is the kernel function and $\varepsilon$ is an  additive independent identically distributed noise expected to be present in the data and is assumed to be a Gaussian with zero mean and variance $\sigma^{2}_{n}$. To allow for this noise, the covariance function is therefore written as
\begin{equation}
\mathrm{cov}_{\eta}(x,y) = k_{\eta}(( x,y) ,( x',y')) + \sigma ^{2}_{n} \delta (( x,y) ,( x',y'))
\end{equation}
where $\displaystyle \delta $ is the Kronecker delta function defined as 
\begin{equation*}
\delta (( x,y) ,( x',y')) =\begin{cases}
1, & x=x'\ \text{and} \ y=y'\\
0, & \ \text{otherwise}
\end{cases}
\end{equation*}

The covariance function is defined using squared exponential kernel $k_{\eta}$ as follows: 
\begin{equation}\label{eqn:expcov}
\begin{aligned}
k_{\eta}(( x,y) ,( x',y')) = & \ \sigma ^{2}_{\eta}\exp (-\frac{(x-x')^{2}}{2w^{2}_{x}} -\frac{(y-y')^{2}}{2w^{2}_{y}} )
\end{aligned}
\end{equation}
where $\displaystyle \sigma _{\eta}$ is the variance of $v_\eta$; $\displaystyle w_{x}$ and $\displaystyle w_{y}$ are the characteristic length-scale parameters, the inference of which is discussed in detail in Section \ref{sec:inference}.

In the training dataset, each observation $s_i=\{(v^i_{xj},v^i_{yj},x^i_j,y^i_j)|j=1,2,...,l_i,(x^i_j,y^i_j)\in A_{ROI}\}$, referred to as a frame, is a sample from the time-series data sequence $\mathcal{S}=\{s_i|i=1,2,...,N\}$ and is a 2-dimensional representation of a given $A_{ROI}$. Each frame contains the position information $(x,y)$ and the corresponding velocity information $v_{\eta}$ of all the vehicles observed in $A_{ROI}$ at that time instance. In vector form, the observed data is given by
\begin{equation}
s =( V_{x} ,V_{y} ,X,Y) = \ \left( 
\left[ \begin{matrix}
v_{x1} \\ \vdots \\ v_{xl} \end{matrix}\right],
\left[ \begin{matrix}
v_{y1} \\ \vdots \\ v_{yl} \end{matrix}\right],
\left[ \begin{matrix}
x_{1} \\ \vdots \\ x_{l} \end{matrix}\right],
\left[ \begin{matrix}
y_{1} \\ \vdots \\ y_{l} \end{matrix}\right]
\right)
\end{equation}
Here $v_{\eta j} = [ V_{\eta}]_{j}$ is the observed velocity at $ (x_j,y_j) = ([ X]_{j} ,[ Y]_{j})$, where $ [ \cdot ]_{j}$ denotes the $\displaystyle j$th element of a vector. Similarly, we write the test frame data as $\displaystyle s^{*} =\left( V^{*}_{x} ,V^{*}_{y} ,X^{*} ,Y^{*}\right)$, where $\displaystyle V^{*}_{\eta}$ is unknown.

The definition of GP indicates that the velocity output is a joint Gaussian distribution \cite{GPforML}, given by
\begin{equation} \label{eqn:JointGaussian}
\begin{split}
\begin{bmatrix}
V^{*}_{\eta}\\
V_{\eta}
\end{bmatrix} &\sim 
\mathcal{N} \left(
\begin{bmatrix}
\mu _{\eta}\left( X^{*} ,Y^{*}\right)\\
\mu _{\eta}\left( X ,Y\right)
\end{bmatrix}, 
\right.
\\ 
&\left.
\begin{bmatrix}
K_{\eta}\left( X^{*} ,Y^{*} ,X^{*} ,Y^{*}\right) & K_{\eta}\left( X^{*} ,Y^{*} ,X,Y\right)\\
K_{\eta}\left( X,Y,X^{*} ,Y^{*}\right) & K_{\eta}\left( X,Y,X,Y\right) + \sigma^{2}_{n} I
\end{bmatrix}
\right)
\end{split}
\end{equation}
where $\mu _{\eta}\left( X^{*} ,Y^{*}\right)$ and $\mu _{\eta}\left( X ,Y\right)$ are the mean vectors treated as the prior distribution of the velocity. If there are $n$ training data points and  $n^*$ test points, then $K_{\eta}\left( X,Y,X^{*} ,Y^{*}\right)$ denotes the $n$ x $n^*$ matrix of the covariances evaluated at all pairs of the training and test points, and similarly for the other entries $K_{\eta}\left( X^{*} ,Y^{*} ,X,Y\right)$, $K_{\eta}\left( X,Y,X,Y\right)$ and $K_{\eta}\left( X^{*} ,Y^{*} ,X^{*} ,Y^{*}\right)$. Conditioned on the observation $\displaystyle V_{\eta}$, the predictive distribution of $V^{*}_{\eta}$ is still a Gaussian
\begin{equation}\label{eqn:GPposterior}
\resizebox{1\hsize}{!}{
  $V^{*}_{\eta} |V_{\eta} ,\left( X^{*} ,Y^{*}\right) ,( X,Y) \sim \mathcal{N}\left( \mu ^{*}_{\eta}\left( X^{*} ,Y^{*}\right) ,\mathrm{cov}^{*}_{\eta}\left( X^{*} ,Y^{*}\right)\right)$}
\end{equation}
where
\begin{equation*}\label{eqn:GPposteriorMeanCov}
  \resizebox{1\hsize}{!}{
  $\begin{array}{ r l }
\mu ^{*}_{\eta}\left( X^{*} ,Y^{*}\right) =& \mu _{\eta}\left( X^{*} ,Y^{*}\right) +K_{\eta}\left( X^{*} ,Y^{*} ,X,Y\right)\\
 & \left[ K_{\eta}( X,Y,X,Y) +\sigma ^{2}_{n} I\right]^{-1}( V_{p} -\mu _{\eta}( X,Y))\\
\\ \mathrm{cov}^{*}_{\eta}\left( X^{*} ,Y^{*}\right) =& K_{\eta}\left( X^{*} ,Y^{*} ,X^{*} ,Y^{*}\right) -K_{\eta}\left( X^{*} ,Y^{*} ,X,Y\right)\\
 & \left[ K_{\eta}( X,Y,X,Y) +\sigma ^{2}_{n} I\right]^{-1} K_{\eta}\left( X,Y,X^{*} ,Y^{*}\right)
\end{array}$}
\end{equation*}

\subsubsection{Multi-vehicle trajectory generation from motion patterns}
In order to calculate the likelihood of a GP motion pattern $g$ given frame  $s = \{( v_{xj} ,v_{yj}, x_{j} ,y_{j}) \ |\ j=1,2,\cdots l,\ ( x_{j} ,y_{j}) \in A_{ROI}\}$, we need to specify how $g$ generates $s$. We model this through a three-step generative procedure by drawing (1) the number of vehicles $\displaystyle l$, (2) the location of all the vehicles $\displaystyle \{( x_{j} ,y_{j}) \}$, and (3) the velocity $\displaystyle \{( v_{xj} ,v_{yj})\}$ of all vehicles.

Let $\displaystyle a_{m} '=\#(\{s_{i} \ |\ l_{i} =m,s_{i} \in \mathcal{S}\})$ be the number of frames with $\displaystyle m$ vehicles observed, $\displaystyle m=1,2,\cdots $, and assign the weights $\displaystyle a_{m} =a'_{m} /\Sigma ^{\infty }_{q=1} a'_{q}$ to $\displaystyle \lambda ( m)$, where $\displaystyle \lambda ( m)$ is the distribution concentrated at a single point $\displaystyle m$. Then, the empirical distribution $\phi (\mathcal{S})$  of the number of vehicles $\displaystyle l$ is as follows
\begin{equation}
l\sim \phi (\mathcal{S}) =\sum ^{\infty }_{q=1} a_{q} \lambda ( q)
\end{equation}

Next, we construct an empirical distribution for the locations $\displaystyle \{( x_{j} ,y_{j}) \}$. We first discretize $ A_{ROI}$ into disjoint bins $\displaystyle \mathcal{A} =\{A_{1} ,A_{2} ,\cdots ,A_{{n}_{A}}\}$, such that $\displaystyle A_{\theta} \cap A_{\Bar{\theta}} =\emptyset $ for any $\displaystyle \theta \neq \Bar{\theta};  \theta ,\Bar{\theta}\in \{1,2,\cdots ,n_{A}\}$ and $\displaystyle \cup _{i} A_{i} =A_{ROI}$. In each bin $A_\theta$, we then account for the number of agents appeared, denoted as $\displaystyle a'_{A_{\theta}} =\#\left(\left\{\left( x^{i}_{j} ,y^{i}_{j}\right) \ |\ \left( x^{i}_{j} ,y^{i}_{j}\right) \in A_{\theta} \cap s_{i} ,\ s_{i} \in \mathcal{S}\right\}\right)$, and assign a weight \ $\displaystyle a_{A_{\theta}} =a'_{A_{\theta}} /\Sigma ^{n_{A}}_{\Bar{q}=1} a'_{A_{\Bar{q}}}$ to bin $\displaystyle A_{\theta}$. We have:
\begin{equation}
( x_{j} ,y_{j}) \sim \psi ( \mathcal{S},\mathcal{A}) =\sum ^{n_{A}}_{\Bar{q}=1} a_{A_{\Bar{q}}} u( A_{\Bar{q}}) ,\ j=1,2,\cdots ,l
\end{equation}
where $\displaystyle u( A_{\Bar{q}})$ is a uniform distribution over bin $\displaystyle A_{\Bar{q}}$. 

To sample the velocity for each vehicle from a given motion pattern, similar to the notation in (\ref{velGP}), we have 
\begin{equation}
V_{\eta} \sim \mathcal{N}( \mu _{\eta}( X ,Y) ,K_{\eta}( X,Y,X,Y)).
\end{equation}
in vector form of the frame data.

Therefore, the likelihood of motion pattern $g$ given frame $s$ is calculated as 
\begin{equation}\label{likelihood}
\begin{split}
p(s|g) =&\phi ( l;\mathcal{S}) \cdot \prod ^{l}_{j=1} \psi (( x_{j} ,y_{j}) ;\mathcal{S},\mathcal{A}) \cdot \\
&\prod _{\eta\in \{x,y\}} \mathcal{N}( V_{\eta} ;\mu _{\eta}( X,Y) ,K_{\eta}( X,Y,X,Y))
\end{split}
\end{equation}

For a given dataset, the empirical distributions $\phi$ and $\psi$ are implicitly defined by the data. However, the discussed formulation enables the model to scale to data generation platforms where the distributions are expected to be explicitly defined.

\subsection{Dirichlet Process Mixture of Motion Model}

\begin{figure}[t]
	\centering
	\includegraphics[width = 0.9\linewidth]{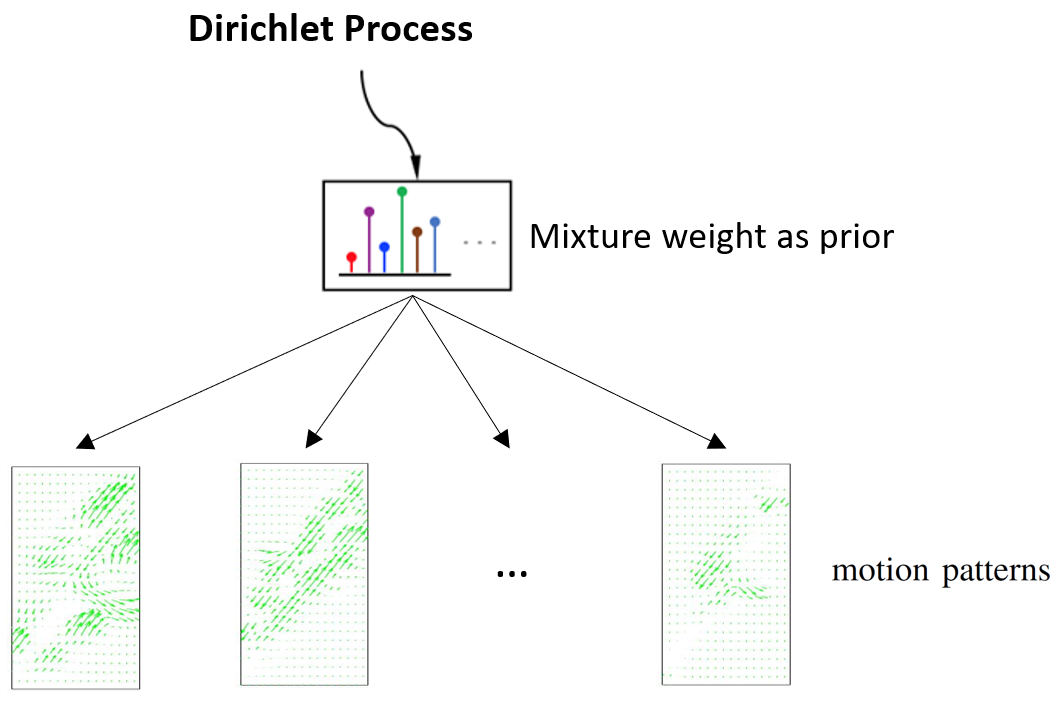}
	\caption{Mixture of motion model with DP prior}
	\label{fig_DP}
\end{figure}

The proposed model considers the dataset $ \mathcal{S}=\{s_{i} \ |\ i=1,2,...,N\}$ as generated by an infinite mixture of motion patterns as shown in (\ref{eqn_Mixture}). Since the total number of the motion patterns $\displaystyle K$ is not known, we give $\displaystyle G$ a Dirichlet Process (DP) prior mixture weight. A DP is a distribution over distributions with infinite components. In our case, however, since the number of observations $N$ is finite, only finitely many components will be discovered from the data.  Fig. \ref{fig_DP} presents the schematic of DP prior as mixture weights for the motion pattern mixture.

An indicator variable $ z_{i}$ is introduced where $ z_{i} =k$ means the frame $s_{i}$ is associated with the latent motion pattern $\displaystyle g_{k}$. The predictive distribution of $g_{z_{i}}$ conditioned on the other motion patterns $g_{{z}_{-i}} =\{g_{z_{k}} \ |\ z_{k} \in z_{-i}\}$, where $\displaystyle z_{-i} =\{z_{k} \ |\ k=1,2,\cdots ,n_k,k\neq i\}$, is
\begin{equation}
g_{z_{i}} \ |\ g_{z_{-i}} ,z_{-i} \sim \frac{1}{\alpha +N-1}\left( \alpha G+\sum _{z_{k} \in z_{-i}} \Delta ( g_{{z}_{k}})\right)
\end{equation}
where $\alpha$ is the concentration parameter  and $ \Delta ( g_{{z}_{k}})$ is the point mass at $\displaystyle g_{{z}_{k}}$. Then the prior probability of $\displaystyle s_{i}$ belonging to an existing motion pattern $ g_{k}$ or an unseen motion pattern $ g_{K+1}$ is given by
\begin{equation}\label{prior}
\begin{aligned}
p(z_{i} =k\ |\ z_{-i} ,\alpha ) & =\frac{n^{-i}_{k}}{N-1+\alpha } ,k = 1,2,\dotsc ,K\\
p(z_{i} =K+1\ |\ z_{-i} ,\alpha ) & =\frac{\alpha }{N-1+\alpha }
\end{aligned}
\end{equation}
where $n_{k}$ is the number of observations currently assigned to $ g_{k}$ and $n^{-i}_{k} =\sum _{z_{\zeta} \in z_{-i}} \mathbf{1}[z_{\zeta} =k]$.  

Combining the likelihood from (\ref{likelihood}) and prior from (\ref{prior}), we have the posterior distribution of $ z_{i}$ as
\begin{equation}\label{eqn:posterior}
\begin{aligned}
p(z_{i} =k\ |\ z_{-i} ,\alpha ,s_{i} ,g_{k} ) =& \ \frac{n^{-i}_{k}}{N-1+\alpha } \ p( s_{i} |g_{k}) ,\\& \quad k= 1,2,\dotsc ,K\\
p(z_{i} =K+1\ |\ z_{-i} ,\alpha ,s_{i}) =& \ \frac{\alpha }{N-1+\alpha } \int\limits_{G} p( s_{i} |g) \ dg
\end{aligned}
\end{equation}
The integration $ \int\limits_{G} p( s_{i} |g) dg$ calculates the likelihood over all the motion patterns contained in the mixture $\displaystyle G$ given observation $\displaystyle s_{i}$.

\begin{figure*}[t]
	\centering
	\includegraphics[width = \linewidth]{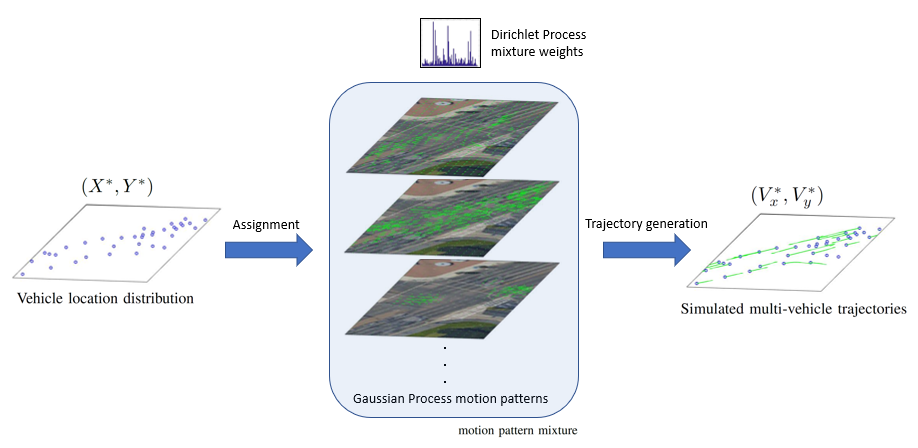}
	\caption{DP-GP mixture model simulation of traffic scenarios}
	\label{fig_DPGP}
\end{figure*}

\section{Model Inference}\label{sec:inference}
In order to find a posterior motion pattern mixture, we use Gibbs sampling to infer the parameters of the model. Every iteration of Gibbs sampling updates the model parameters and the mixture assignment of frames into motion patterns.

\subsection{Mixture Model Assignment}\label{Infer_MixtModel}
The assignment of all frames from $\mathcal{S}$ is performed according to (\ref{eqn:posterior}).

The likelihood $p( s_{i}|g_{k})$ for assigning frame $s_{i}$ into existing pattern $g_{k}$ defined in (\ref{likelihood}) is computed using the GP posterior from (\ref{eqn:GPposterior}), with the training data now given by $s^{g_{k}}=( {V_{x}}^{g_{k}} ,{V_{y}}^{g_{k}} ,X^{g_{k}},Y^{g_{k}})$, which is the vector form of the data of the $n^{-i}_{k}$ frames clustered under $g_{k}$ and the testing data $({V_{x}}^{s_{i}} ,{V_{y}}^{s_{i}} ,X^{s_{i}},Y^{s_{i}})$ of frame $s_{i}$
\begin{equation}
\begin{aligned}
  V^{s_{i}}_{\eta} |{V_{\eta}}^{g_{k}} ,\left( X^{s_{i}} ,Y^{s_{i}}\right) , &( X^{g_{k}},Y^{g_{k}}) \sim 
  \\& \mathcal{N}\left( \mu ^{*}_{\eta}\left( X^{s_{i}} ,Y^{s_{i}}\right) ,\mathrm{cov}^{*}_{\eta}\left( X^{s_{i}} ,Y^{s_{i}}\right)\right)
\end{aligned}
\end{equation}
where $\mu ^{*}_{\eta}$ and  $\mathrm{cov}^{*}_{\eta}$ hold the same definition as in (\ref{eqn:JointGaussian}). A maximum a posteriori estimation is then performed across all the motion patterns $g_{k}$ to identify the assignment $z_{i}$

For assignment of frame $s_{i}$ under a new, unseen pattern $g_{K+1}$, we use Monte-Carlo (MC) integration to approximate the likelihood integral $\int\limits_{G} p( s_{i}|g) dg$. Each MC iteration samples a new motion pattern using priors of model parameters and computes the likelihood using the GP prior given as  
\begin{equation}
V^{s}_{\eta} |\left( X^{s} ,Y^{s}\right) \sim \mathcal{GP}\left(\mu_{\eta0}(X^s,Y^s),K_{\eta0}(X^{s} ,Y^{s},X^{s} ,Y^{s})\right)
\end{equation}
with kernel function 
\begin{equation}
k_{\eta0} (x,y,x',y') = \sigma ^{2}_{\eta0}\exp \left(-\frac{(x-x')^{2}}{2w^{2}_{x0}} -\frac{(y-y')^{2}}{2w^{2}_{y0}}\right)
\end{equation}
where $\mu_{\eta0}$ and $\sigma ^{2}_{\eta0}$ are set to the data mean and variance respectively, and $w_{\eta0}$ is sampled using the prior defined later in (\ref{eqn:w_prior})

\subsection{Model Parameters}

The length scale parameters $w_{x}$ and $w_{y}$ from the exponential covariance calculation in (\ref{eqn:expcov}) are given gamma prior  
\begin{equation}\label{eqn:w_prior}
    w_{\eta} \sim \Gamma(a,b)
\end{equation}
where shape factor $a$ and scale factor $b$ are constants. The posterior calculation of $w_{\eta}$ uses the likelihood given by the GP prior of the frame data $s^{g_{k}}$ assigned under motion pattern $g_{k}$. The parameters $w_{\eta}$ are therefore updated by re-sampling from the posterior given by
\begin{equation}
    w_{\eta}|g_{k} \sim \Gamma(a,b)\cdot p\left(V^{s^{g_{k}}}_{\eta} |( X^{s^{g_{k}}} ,Y^{s^{g_{k}}})\right)
\end{equation}

For the concentration parameter $\alpha $, similar to \cite{rasmussen2000}, an inverse gamma prior is chosen and is updated by re-sampling from the posterior distribution given by
\begin{equation}
p(\alpha |K,N) \propto \frac{\alpha ^{K-3/2}\exp( -1/( 2\alpha )) \Gamma ( \alpha )}{\Gamma ( N+\alpha )}
\end{equation}

The inference algorithm is summarized in Algorithm \ref{algo:inference}.

\begin{algorithm}
\caption{DP-GP Inference\label{algo:inference}}
\textbf{Initialization} \\
$K \longleftarrow 1$\; \\
$w_{\eta} \sim \Gamma(a,b)$ \\
$\alpha^{-1} \sim \Gamma(1,1)$ \\
\For{Gibbs sampling iterations}{
  \SetKwFunction{FMain}{}
  \SetKwProg{Fn}{Update Mixture Assignment}{:}{}
  \Fn{\FMain{}}
{
\For{frames i = 1,2,...,N}
{
\For{motion patterns k = 1,2,...,K}
{
$p(z_{i} =k| z_{-i} ,\alpha ,s_{i} ,g_{k} ) = \frac{n^{-i}_{k}}{N-1+\alpha } \ p( s_{i} |g_{k})$\\
}
$p(z_{i} =K+1| z_{-i} ,\alpha ,s_{i} ) = \frac{\alpha}{N-1+\alpha } \int\limits_{G} p( s_{i} |g)dg$\\
{$Update$} \resizebox{0.83\hsize}{!}{
$z_{i} = \argmax_{k^{+}}p(z_{i} =k^{{+}}| z_{-i} ,\alpha ,s_{i} ,g_{k^+} )$}\\
\ where $k^{+}=1,2,...,K+1$\\
\ {$Update \; K,n_{k}$}
}
}
  \SetKwFunction{FMain}{}
  \SetKwProg{Fn}{Update Model Parameters}{:}{}
  \Fn{\FMain{}}
{
\For{motion patterns k = 1,2,...,K}
{
$w_{\eta} \sim w_{\eta}|g_{k};$
}
$\alpha \sim \alpha |K,N$
}
}
\end{algorithm}
After the Gibbs sampling iterations, the posterior mixture model is used to generate simulated trajectories as shown in Fig. \ref{fig_DPGP}. Given a test frame, a motion pattern assignment is performed using this mixture to generate the GP posterior mean velocity field. The velocity field defines the multi-vehicle trajectory simulation on the test frame.

\section{Experiment And Results}\label{sec:experiments}
\subsection{Experiment Setup}
For evaluating the proposed motion model, we choose a real-world traffic dataset collected as part of Federal Highway Administration's (FWHA) Next Generation SIMulation (NGSIM) project~\cite{NGSIM_highway,NGSIM_intersection}, providing detailed multi-vehicle trajectory data as a time-series sequence. The velocity information as $v_{x}$ and $v_{y}$ components is derived from this trajectory data. The model is evaluated on two traffic settings- highway dataset collected on a segment of the US Highway 101 (Hollywood Freeway) in Los Angeles, and intersection dataset collected on Lankershim Boulevard at Universal Hollywood Dr. in Los Angeles.

The inference algorithm runs for 100 Gibbs sampling iterations and is executed using parallel computing on a 44-core processor. The parameters $a=10$ and $b=1$ are chosen for gamma prior of the length scale parameters. The variance $\sigma_{n}^2$ for the additive Gaussian noise $\varepsilon$ is set to 1.

\subsection{Highway Traffic Scenarios}
\begin{figure}[h]
\centering
\subfigure[]{%
\label{fig_mixtHighway}%
\includegraphics[height = 2.2in]{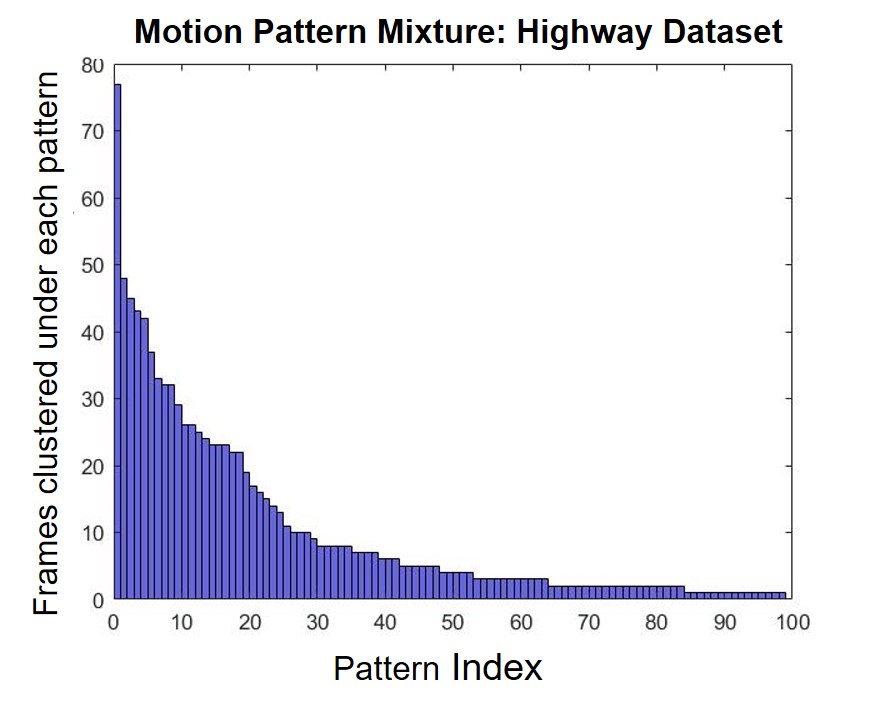}}\\
\subfigure[]{%
\label{fig_mixtIntersection}%
\includegraphics[height = 2.2in]{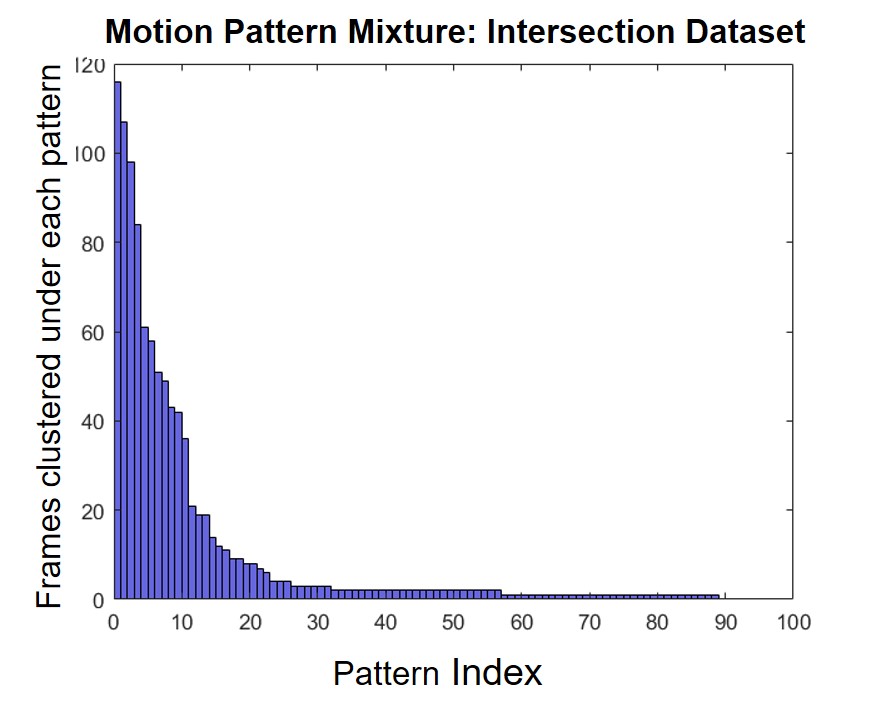}}%

\caption{DP-GP mixture proportion from:
(a) Highway dataset; 
(b) Intersection dataset}
\end{figure}

The highway dataset is down-sampled to 1000 frames of time-sequence data with discretization of 0.5s. The mixture model resulted in 99 motion patterns being extracted from the data. The mixture proportion in a decreasing order is presented in Fig. \ref{fig_mixtHighway}.

\begin{figure}[h]
\centering
\subfigure[]{%
\label{fig:highway_a}%
\includegraphics[height = 2in]{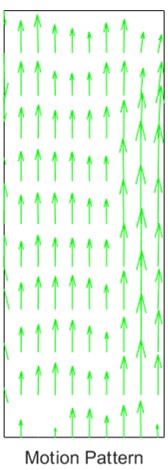}}\hspace{2mm}%
\subfigure[]{%
\label{fig:highway_b}%
\includegraphics[height = 2in]{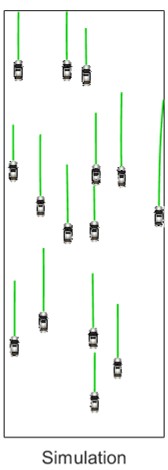}}\hspace{2mm}%
\subfigure[]{%
\label{fig:highway_c}%
\includegraphics[height = 2in]{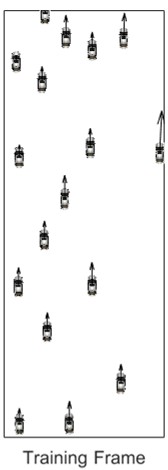}}\hspace{2mm}%
\subfigure[]{%
\label{fig:highway_d}%
\includegraphics[height = 2in]{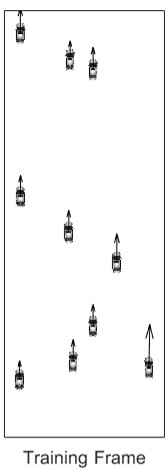}}%
\caption{Highway dataset:
(a) Mean GP velocity field of motion pattern 1; 
(b) Motion pattern 1 based multi-vehicle trajectory simulation of test frame; 
(c) Frame 512: observation clustered under motion pattern 1; 
(d) Frame 894: observation clustered under motion pattern 1}
\label{fig:highway_results}
\end{figure}

We randomly choose a test frame from the dataset to generate the simulated multi-vehicle trajectories from the motion pattern results obtained at the end of the Gibbs sampling iterations. A motion pattern from the mixture is then assigned to the frame according to the assignment procedure discussed in Section \ref{sec:inference}. The derived mean GP velocity field is imposed on the vehicle distribution present in the $A_{ROI}$ to run the simulation for that interaction scenario, and is presented in Fig. \ref{fig:highway_results}. To illustrate the clustering, the original observations from the data that were assigned under the same motion pattern are also included (with the vehicle velocity vectors shown) in the figure.

\subsection{Intersection Traffic Scenarios}
We reproduce the results of the highway dataset for the intersection dataset, which is down-sampled to 1000 frames of time-sequence data with discretization of 0.5 s. The posterior mixture model consists of 89 motion patterns whose mixture proportion in a decreasing order is presented in Fig. \ref{fig_mixtIntersection}.

\begin{figure}[h]
\centering
\subfigure[]{%
\label{fig:intersection_map}%
\includegraphics[height = 2in]{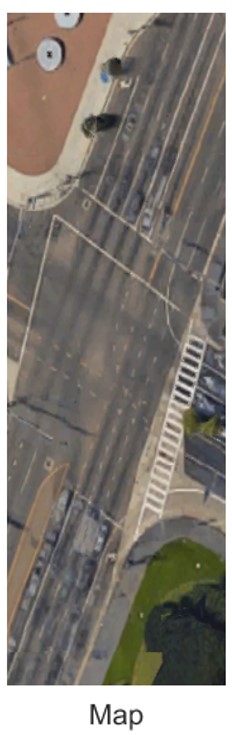}}\hspace{1mm}%
\subfigure[]{%
\label{fig:intersection_a}%
\includegraphics[height=2in]{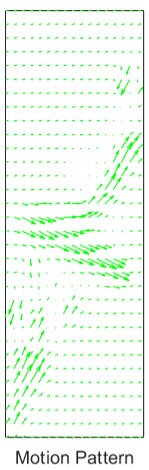}}\hspace{1mm}%
\subfigure[]{%
\label{fig:intersection_b}%
\includegraphics[height=2in]{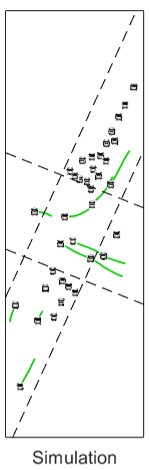}}\hspace{1mm}%
\subfigure[]{%
\label{fig:intersection_c}%
\includegraphics[height=2in]{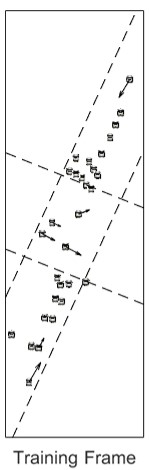}}\hspace{1mm}%
\subfigure[]{%
\label{fig:intersection_d}%
\includegraphics[height=2in]{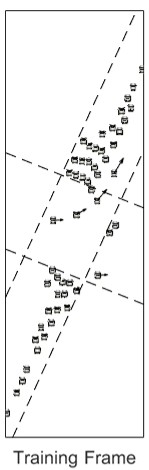}}\hspace{1mm}%
\caption{Intersection dataset: 
(a) Intersection image; 
(b) Mean GP velocity field of motion pattern 1; 
(c) Motion pattern 1 based multi-vehicle trajectory simulation of test frame; 
(d) Frame 211: observation clustered under motion pattern 1; 
(e) Frame 310: observation clustered under motion pattern 1}
\label{fig:intersection_results}
\end{figure}

Fig. \ref{fig:intersection_results} shows a simulated trajectory with the motion pattern vector field and the data observations similar to the results seen from the highway dataset.

\begin{figure}[h]
\centering
\subfigure[]{%
\label{fig:intersection_patterna}%
\includegraphics[height = 2in]{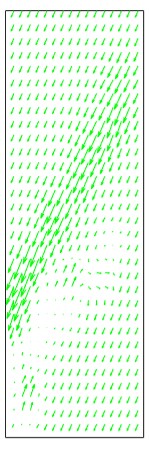}}\hspace{2mm}%
\subfigure[]{%
\label{fig:intersection_patternb}%
\includegraphics[height = 2in]{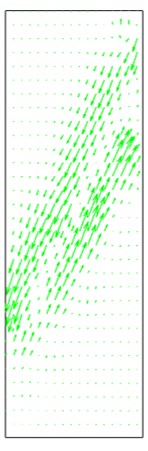}}\hspace{2mm}%
\subfigure[]{%
\label{fig:intersection_patternc}%
\includegraphics[height = 2in]{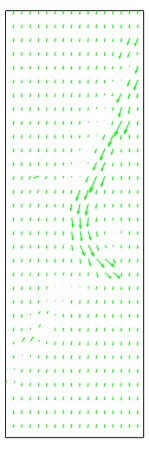}}\hspace{2mm}%
\subfigure[]{%
\label{fig:intersection_patternd}%
\includegraphics[height = 2in]{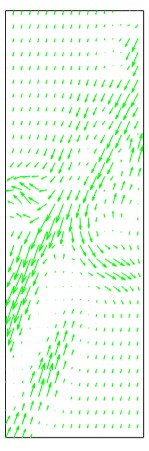}}%
\caption{Intersection dataset: Mean GP velocity field of 
(a) motion pattern 5; 
(b) motion pattern 19; 
(c) motion pattern 64; 
(d) motion pattern 85}
\label{fig:intersection_patterns}
\end{figure}

\section{Discussion}\label{sec:discussion}

\subsection{Result Analysis}
The results from both the datasets demonstrate that the proposed model when applied to large time-series data sequences, extracts the underlying motion patterns which can be used to represent the interaction scenarios. 

We would like to note in the case of both the datasets that the frame indices of the observation (training) frames clustered under the presented motion patterns are far apart in the data sequence. In the highway dataset, the frame indices in \ref{fig:highway_c} and \ref{fig:highway_d} show that these observations are over 400 frames i.e. over 200 s apart. The frame indices from the intersection results in Fig. \ref{fig:intersection_results} also show similar effect.

The motion pattern from highway dataset presented in Fig. \ref{fig:highway_results} captures a scenario with vehicles in the rightmost lane of the $A_{ROI}$ making an exit from the highway. This is learnt from the observations which include \ref{fig:highway_c} and \ref{fig:highway_d}. We also see the same reflected in the simulated trajectory \ref{fig:highway_b}. 

For the intersection based $A_{ROI}$, the GP mean velocity field of the generated motion pattern presented in Fig. \ref{fig:intersection_a} and the corresponding simulated trajectory \ref{fig:intersection_b} indicate a motion scenario where amongst the vehicles incoming from one direction, some are seen to take a left turn at the intersection while others continue straight with vehicles from the other directions standing still. This is a scene that the model generated on the test frame after learning the motion pattern from the data. The GP mean velocity field and simulated trajectory results show that the model learns the road's physical layout from the data by exhibiting almost non-existent probability of the posterior vector field outside the road boundary. We also see the model pick up on lane information such as which lanes correspond to a left turn, without having any explicit information of the road layout. The same effect is retained in the results presented in Fig. \ref{fig:intersection_patterns}, as further discussed here.

While the highway dataset has been clustered into motion patterns as expected, it offers little insight into the semantic visualization of the results due to the vehicle interactions being limited to motion in only one direction. In that sense, the intersection results offer better diversity based on the interactions involving vehicles' motion in multiple directions. Illustrating this, other motion patterns generated from the intersection dataset are presented in Fig. \ref{fig:intersection_patterns}. While the mean velocity field of pattern \ref{fig:intersection_patterna} captures the motion of vehicles at the intersection in one direction with the vehicles from oncoming side at standstill, \ref{fig:intersection_patternb} presents the interaction scenario where vehicles from both sides travel straight. \ref{fig:intersection_patternc} almost exclusively captures the left turning motion of the vehicles from one direction and \ref{fig:intersection_patternd} presents a more complex scenario with vehicle motion flow in many directions. 


\subsection{Limitations}
The primary limitation of this work lies in the inference of the mixture model using Gibbs sampling. A termination criterion is not explicitly available, especially because of the unsupervised nature of the problem, which makes it difficult to come up with a suitable number of iterations. An evaluation of the resultant mixture assignment could be defined to determine the convergence. 

Furthermore, in a dataset involving a more complex intersection, modeling all the multi-vehicle interactions present within a given frame using a single Gaussian process might overly marginalize the true velocity information of the data. Future research might employ multiple Gaussian processes possibly conditional on vehicle direction of motion and other information to model each interaction scenario. 

There could exist motion patterns with the training frame data not extending to all the parts of the region of interest. Because the Gaussian process modeling spans across the entire position domain, the resulting velocity field could generate simulations with unreasonable vehicle trajectories in the parts of the region of interest with limited data. Also, the model does not consider the true road layout information. Although the results show that the model implicitly learns this from the data, the simulated trajectories can be treated with greater confidence if the road boundary and other traffic rules-based information are embedded into the model.
\section{Conclusion and Future Work}\label{sec:conclusion}
In this work, we formulate a model for multi-vehicle interaction scenarios using GP, a mixture of which is generated from naturalistic data by using non-parametric Bayesian learning. By employing DP prior for the mixture model assignment, we alleviate the restriction on the number of motion patterns existing in the dataset, allowing the model to be fully data-driven. The experiment results using NGSIM datasets demonstrate the extracted multi-vehicle interactions as motion patterns, capable of capturing the highly dynamic scenes from highways and intersections. This result allows modelers to extract multi-vehicle interaction scenarios efficiently from large-scale data, which can be further used for simulating traffic scenes, predicting the trajectories of vehicles in multi-vehicle systems, and efficiently evaluating the safety of an AV when interacting with human driven vehicles or other AVs in complex driving situations.

The code for this work is available at \url{https://github.com/zhao-lab/kalidindi_dpgp_multi_vehicle_2019}


\section*{Acknowledgment}

Toyota  Research  Institute  (“TRI”)  provided  funds  to  assist the  authors  with  their  research  but  this  article  solely  reflects the opinions and conclusions of its authors and not TRI or any other Toyota entity.

\bibliographystyle{IEEEtran}
\bibliography{References.bib}

\end{document}